\documentclass[english,runningheads]{lni}
\usepackage{booktabs}

\usepackage[]{blindtext}

\usepackage{tikz}
\usetikzlibrary{positioning,calc,arrows.meta}

\newcommand{\code}[1]{\nolinkurl{#1}}

\newif\ifIEEEpaper
\IEEEpaperfalse
\newcommand{\paperscopeheading}{\paragraph{Scope and limitations.}}
\newcommand{\paperfigurename}{Figure}
\newcommand{\scatterwidth}{0.55\textwidth}
\newcommand{\latencywidth}{0.78\textwidth}
\newenvironment{paperwidefigure}{\begin{figure}[t]}{\end{figure}}
\newenvironment{paperwidetable}{\begin{table}[t]}{\end{table}}

\begin{document}
\pagenumbering{gobble}
\hypersetup{pageanchor=false}
\title[Not All Challenges Are Equal]{Not All Challenges Are Equal: Evaluating Interactive Liveness Detection Against Real-Time Face Swaps}

\author[1]{Anton Firc}{ifirc@fit.vut.cz}{0000-0002-4717-1910}
\author[1]{David Drtil}{}{}
\author[1]{Martin Perešíni}{iperesini@fit.vut.cz}{0000-0002-2875-9567}
\author[1]{Milan Šalko}{isalko@fit.vut.cz}{0009-0004-9604-168X}%
\author[1]{Vojtěch Staněk}{istanek@fit.vut.cz}{0009-0000-5722-0571}%
\author[1]{Kamil Malinka}{malinka@fit.vut.cz}{0000-0002-9009-2193}%
\affil[1]{Security@FIT, Brno University of Technology, Brno, The Czech Republic}

\editor{\small F.  Boutros,  N. Damer,  M. Fang,  M. Gomez-Barrero, P. Neto, K. Raja,  \linebreak C. Rathgeb, A. Sequeira, M. Todisco and R. Tolosana:} 
\booktitle{BIOSIG 2026} 
\maketitle

\begin{abstract}
Remote identity verification assumes that a camera stream shows a live person, yet public real-time face-swap tools can replace that person before the stream reaches the verifier. Passive deepfake detectors generalize poorly to new generators, while interactive challenges are often selected without evidence of which gestures expose current attacks. We evaluate randomized gesture challenges in 108 live sessions with six subjects, 24 impersonated identities, and two face-swap tools, \emph{DeepFaceLive} and \emph{FaceFusion}, injected through a virtual camera. The static calibration segment reaches a session-level AUC of 0.878 with a frozen UCF spatial detector, while the mean across the active-gesture windows reaches 0.988. Pose and combined gestures have the highest point estimates, whereas expressions show no clear improvement. Hand occlusion against \emph{FaceFusion}, configured with an occlusion mask, scores below the static baseline. The end-to-end time required to satisfy the validator also distinguishes attacks from genuine sessions, achieving a session-level AUC of 0.732, although this signal includes synthesis and transport delays. These results show that challenge sets should be tested against current attack configurations and that response time can provide an additional, deployment-specific signal.
\end{abstract}


\section{Introduction}
\label{sec:Introduction}

Remote identity verification, from video calls to online banking onboarding, rests on a strong premise: that the camera shows a real person, in real time, in front of a real lens. Deep learning has reduced the cost of a convincing synthetic face to minutes of setup~\cite{01:Firc:DeepfakesThreat:2023}. A single public photo feeds a real-time face-swap pipeline on a consumer graphics card, and the attacker routes the synthesized stream into the call through a virtual camera driver. No server compromise, network tampering, or physical artifact is needed. The threat is real: a widely reported case includes a 25 million USD transfer authorized after a video call with a deepfake CFO~\cite{67:CNNDeepfakeCFO2024}.

Traditional passive detection searches individual frames for digital inconsistencies such as blending seams or unnatural textures. It catches older generators but generalizes poorly across the moving target of new tools~\cite{12:Pei:GenAndDetBenchmark:2024}, and deployed liveness systems fall to injected streams even when the vendor advertises an active test~\cite{03:Wang:LivingALie:2024}. A static detector alone is not a reliable strategy for a service that must keep verifying users when the next generator releases.

A natural response is to add an interactive layer to the passive pipeline. The user performs a short, randomized sequence of actions, such as rotating the head, changing expression, or covering part of the face, within a fixed time. These actions can expose synthesis errors when the swap must reproduce unusual poses or occlusions in real time. Randomization reduces the risk of simple pre-recorded replay, although it does not prevent an attacker from generating and injecting a live response. Such challenges are widely deployed in commercial liveness verification~\cite{04:Li:SeeingIsLiving:2022, 03:Wang:LivingALie:2024}, but their selection is often based on intuition. This proof-of-concept study measures which gesture categories improve detection relative to current real-time face swaps and identifies cases in which a challenge provides little or no benefit.

We implement a hybrid architecture that combines frozen passive detectors with randomized gesture challenges and evaluate it in 108 live sessions with six subjects, 24 impersonated identities, and two real-time face-swap tools, \emph{DeepFaceLive}~\cite{13:Liu:DeepFaceLab:2023} and \emph{FaceFusion}~\cite{64:Ruhs:FaceFusion:2023}, injected through a virtual camera. Using the raw UCF~\cite{28:Yan:UCF:2023} spatial score without fusion weights, the static calibration segment reaches a session-level AUC of 0.878, while the mean across the active-gesture windows reaches 0.988. The improvement is not uniform: pose and combined gestures have the highest point estimates, expressions show no clear gain, and occlusion against \emph{FaceFusion} scores below the static baseline. Because \emph{FaceFusion} was tested only with its occlusion mask enabled, this result is consistent with a tool-specific countermeasure but does not isolate its causal effect. The campaign also revealed that the tested attack pipelines take longer to satisfy the validator. This end-to-end timing signal reaches a session-level AUC of 0.732, but it includes synthesis and virtual-camera delay as well as user response time. The work described in this paper results from a previously completed master`s thesis~\cite{drtil2026deepfake}.

In summary, the main outcomes of this work are:

\begin{itemize}
    \item An end-to-end liveness pipeline combining frozen passive detection with randomized gesture challenges, evaluated in 108 live genuine and attack sessions.
    \item A subject-clustered comparison of gesture categories against a static baseline, showing aggregate improvement but different effects across attack configurations.
    \item An empirical evaluation of verifier-observed time-to-validation as a complementary end-to-end signal during live face-swap attacks.
\end{itemize}

\section{Background and Related Work}
\label{sec:RelatedWork}

\textbf{Passive detection and its generalization gap.} Passive detection analyzes video without requiring additional user action; its cues range from synthesis artifacts to physiological inconsistencies, such as the spatial and temporal biological signals used by \emph{FakeCatcher}~\cite{68:Ciftci:FakeCatcher:2020}. Artifact-based detectors can learn cues associated with the manipulations present in their training data, and their performance can decline when tested on different manipulation methods or data distributions. Cross-dataset evaluations and an in-the-wild benchmark of deepfakes circulated in 2024 document this generalization gap~\cite{12:Pei:GenAndDetBenchmark:2024,19:Chandra:DeepfakeEval:2025}. Methods based on self-blended training images~\cite{06:Shiohara:SelfBlended:2022}, common forgery features~\cite{28:Yan:UCF:2023}, and pretrained vision-language features~\cite{30:Ojha:UniversalFake:2023} aim to improve transfer beyond the manipulations used for training. The deployed pipeline uses \emph{UCF}~\cite{28:Yan:UCF:2023} and \emph{CViT2}~\cite{51:Wodajo:CViT2:2024} with frozen pretrained weights, while the evaluation below reports raw \emph{UCF} scores on live \emph{DeepFaceLive} and \emph{FaceFusion} outputs without generator-specific adaptation.

\textbf{Real-time face-swap tools.} Mature open-source toolchains make live face swapping accessible. \emph{DeepFaceLab}~\cite{13:Liu:DeepFaceLab:2023} trains a model for a specific identity pair, while \emph{DeepFaceLive} applies the exported \texttt{.dfm} model to a webcam stream. Community models for public figures can be downloaded without training a new model. \emph{FaceFusion}~\cite{64:Ruhs:FaceFusion:2023} instead swaps identities from reference images using \emph{InsightFace} embeddings~\cite{62:Guo:InsightFace:2018}. Its optional \emph{XSeg} occlusion mask preserves foreground objects, such as hands, by excluding them from the swapped face region. The attacker can route either tool through a virtual camera into an application that reads the operating-system camera API~\cite{03:Wang:LivingALie:2024}. Residual artifacts include flicker, mask-boundary errors, identity drift at extreme poses, and failures around occluders; interactive challenges are intended to expose these weaknesses.

\textbf{Interactive liveness detection.} Interactive liveness asks the user to perform an action within a time window and verifies the action in the video. Common examples include head rotation, blinking measured through the Eye Aspect Ratio~\cite{21:Soukupova:RealTimeEyeBlinking:2016}, expressions measured through landmarks or dedicated networks~\cite{52:Ming:FaceLiveNet:2018}, and hand-over-face occlusion. A pre-rendered deepfake cannot answer an arbitrary instruction, but a real-time swap can transfer the motion of its human driver~\cite{04:Li:SeeingIsLiving:2022}. Motion-only checks are therefore insufficient, as demonstrated against commercial services~\cite{04:Li:SeeingIsLiving:2022} and mobile SDKs~\cite{03:Wang:LivingALie:2024}. Hybrid designs instead use the challenge to provoke synthesis artifacts that a passive detector can measure~\cite{02:Firc:FaceAuth:2024}. \emph{DF-Captcha} proposed this challenge-response principle~\cite{66:Mirsky:DFCaptcha:2022}, while \emph{GOTCHA} evaluated visible degradation on recordings from 47 users and 2022-era pipelines~\cite{65:Mittal:GOTCHA:2024}. We extend this evidence with live end-to-end sessions, subject-clustered estimates of category-level effects against current attack configurations, and an empirical analysis of verifier-observed time-to-validation.

\section{Proposed Verification Pipeline}
\label{sec:Method}

The pipeline is a thin-client design: the browser only captures the webcam stream and displays instruction overlays, while all analysis runs on the server. A central controller manages the session as a state machine, drawing random actions from a challenge generator and enforcing a per-action timeout. Two branches analyze the incoming frames in parallel: a passive branch scoring every frame for deepfake artifacts and an interactive branch validating the requested action, and a decision logic module fuses both into the session verdict. \paperfigurename~\ref{fig:architecture} shows the structure.

\begin{paperwidefigure}
\centering
\begin{tikzpicture}[
    font=\scriptsize,
    box/.style={draw=black!60, fill=black!4, rounded corners=1pt, align=center, inner sep=3pt, minimum height=8mm},
    boxP/.style={box, draw=blue!60!black, fill=blue!8},
    boxI/.style={box, draw=orange!80!black, fill=orange!12},
    arr/.style={-{Stealth[length=1.6mm]}, semithick, black!70}]
  \node[box] (client) {Browser client\\camera + overlay};
  \node[box, right=7mm of client] (pre) {Pre-processing\\detect, align, 15\,FPS};
  \node[boxP, anchor=west] (passive) at ($(pre.east)+(0.7,0.65)$) {Passive branch\\UCF, CViT2, identity};
  \node[boxI, anchor=west] (inter) at ($(pre.east)+(0.7,-0.65)$) {Interactive branch\\action validators};
  \node[box, anchor=west] (dec) at ($(pre.east)+(4.0,0)$) {Decision\\logic};
  \node[box, right=6mm of dec] (verdict) {Verdict\\genuine / attack};
  \node[boxI, below=5.5mm of pre] (gen) {Challenge generator + timer};
  \draw[arr] (client) -- (pre);
  \draw[arr] (pre.east) -- (passive.west);
  \draw[arr] (pre.east) -- (inter.west);
  \draw[arr, blue!60!black] (passive.east) -- (dec.west);
  \draw[arr, orange!80!black] (inter.east) -- (dec.west);
  \draw[arr] (dec) -- (verdict);
  \draw[arr, orange!80!black] (gen.east) -| (inter.south);
  \draw[arr, dashed, orange!80!black] (gen.west) -| (client.south) node[pos=0.25, below, font=\tiny]{instructions};
\end{tikzpicture}
\caption{The verification pipeline: a passive branch (blue) scores frames for deepfake artifacts while an interactive branch (orange) validates randomized, time-bounded challenges; the decision logic fuses both.}
\label{fig:architecture}
\end{paperwidefigure}

\textbf{Pre-processing.} Frames are processed at approximately 15\,FPS through the \emph{InsightFace} \texttt{buffalo\_l} pack~\cite{62:Guo:InsightFace:2018} (face detection, landmark regression, \emph{ArcFace} head~\cite{25:Deng:ArcFace:2019}). Landmarks are stabilized with a One Euro filter~\cite{53:Casiez:OneEuro:2012} and an affine alignment produces a canonical crop per detector. The pipeline runs with a P95 per-frame latency below 40\,ms on a single consumer GPU.

\textbf{Passive branch.} Two detectors with frozen public weights score each aligned crop: \emph{UCF}~\cite{28:Yan:UCF:2023} produces a spatial score per frame, while \emph{CViT2}~\cite{51:Wodajo:CViT2:2024} predictions are averaged over sliding frame windows. The models were not trained on outputs from \emph{DeepFaceLive} or \emph{FaceFusion}. An additional identity tracker anchors an \emph{ArcFace} embedding on the first valid frame and monitors cosine similarity throughout the session. The deployed verdict uses these signals, but the gesture comparisons reported below use only the raw UCF spatial score.

\textbf{Interactive branch.} Action validation runs on the \emph{MediaPipe} Face Mesh and Hands models~\cite{37:Lugaresi:MediaPipe:2019} (468 face landmarks with blendshapes, 21 per hand). The gesture pool covers four categories: \emph{pose} (head rotations and leans, verified through Euler angles), \emph{expression} (smile, open mouth, eyebrows, verified through landmark distances and blendshapes), \emph{occlusion} (covering an eye, the nose, or the mouth, verified by intersecting a hand mask~\cite{40:Sanchez:HandSegmentation:2023} with the facial region), and \emph{complex} (two constraints in one hold, e.g., covering the mouth while turning left). A session opens with a two-second \emph{Hold Still} calibration, followed by a randomized sequence of active actions, each bounded by a timeout; the validator reports success or failure for each action.

\textbf{Decision logic.} The deployed system requires every challenge to be completed in time, fuses calibrated passive scores with per-category weights, and flags a session irrecoverably on a single high-confidence artifact detection. These engineering choices are not the subject of this paper: the analysis in Section~\ref{sec:Results} bypasses the fusion and works on raw per-gesture detector responses, so the reported ranking is independent of any weighting.

\section{Experimental Methodology}
\label{sec:Evaluation}

\textbf{Setup and participants.} All sessions ran on a single server (NVIDIA RTX~5080, AMD Ryzen~9, 64\,GB RAM), with webcam capture pinned to 1280$\times$720 at 30\,FPS via an exact \texttt{getUserMedia} constraint. Recording took place in three rooms under diffuse daylight, subjects seated 50--70\,cm from the camera. Six volunteers participated under written consent: three men (25, 27, 51) and three women (24, 24, 49), all from the same broad ethnic group, appearing in the data only under anonymized identifiers. Each subject recorded 6 genuine and 12 attack sessions, for a total of 108 sessions (36 genuine, 36 \emph{DeepFaceLive}, 36 \emph{FaceFusion}), with roles shuffled to suppress learning effects. All decision thresholds were calibrated before the campaign on recordings of the author only. Each retained session begins with a two-second \emph{Hold Still} calibration followed by 5, 6, or 9 active actions (22, 77, and 9 sessions, respectively; median six). Across the campaign, these actions span pose, expression, occlusion, and combined or sequential challenges, with their order randomized within each session. Labels are assigned at recording time.

\textbf{Attack generation.} Attack sessions inject a real-time face swap through a virtual camera. \emph{DeepFaceLive} runs pre-trained community \texttt{.dfm} models, two identities per subject with three sessions each. \emph{FaceFusion} runs \texttt{inswapper\_128} with the \texttt{xseg\_1} occlusion module and four reference photos per identity (two frontal, two slight profiles). Attack identities are matched to the subject on gender and ethnic group, the two attributes that most reliably break a swap when mismatched, with an \emph{ArcFace} cosine ranking as a tie-breaker; the campaign covers 24 attack identities in total.

\textbf{Analysis protocol.} The action window, comprising the frames for which one requested action is held, is the gesture-level unit of analysis. Gesture and session AUCs use the mean raw UCF spatial score and bypass the deployed fusion weights and calibrations. We report ROC AUC between attack and genuine windows by gesture category and generator. Confidence intervals use 2\,000 subject-clustered bootstrap resamples, with comparisons against the \emph{Hold Still} baseline paired within each resample. Because the study contains only six subject clusters and examines several categories--generator contrasts, these intervals are descriptive and unadjusted for multiplicity. Session scores compare the mean UCF response from the single initial calibration segment with the mean across all subsequent active-gesture windows, thereby evaluating active capture as a complete protocol. Challenge-response latency is the interval from the end of one action's hold window to the beginning of the next, converted through the measured session frame rate. Attack sessions include synthesis and virtual-camera transport that are absent from genuine sessions, so latency reflects the complete attack condition rather than an isolated human response time. Timed-out sessions were excluded during collection, but their number and durations were not retained; we therefore do not treat the observed latency difference as a quantified lower bound. We additionally report APCER and BPCER for the deployed system following ISO/IEC 30107-3~\cite{58:ISO:30107-3:2023}, while noting that virtual-camera injection falls outside the standard's presentation-attack scope.

\paperscopeheading
This proof-of-concept study includes six subjects from one broad ethnic group and two real-time face-swap tools. Accordingly, the reported performance and gesture-level effects should be interpreted within the tested population and attack configurations. The latency results capture the end-to-end overhead of the tested attack pipelines and may vary across implementations and hardware.

\section{Results}
\label{sec:Results}

\subsection{Contribution of the Interactive Layer}
\label{subsec:EndToEnd}

\begin{figure}[t]
    \centering
    \includegraphics[width=\scatterwidth]{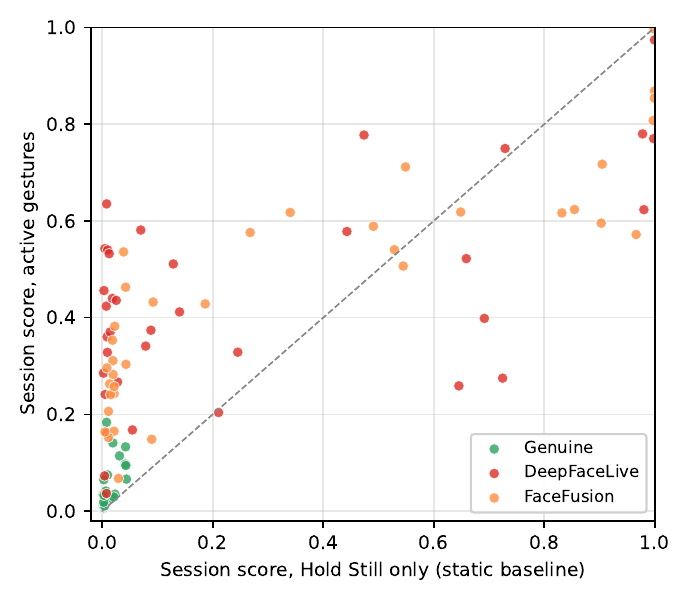}
    \caption{Per-session raw spatial score on the \emph{Hold Still} baseline (x) versus the mean across the active-gesture windows (y). Points above the diagonal have a higher score during active capture than during calibration.}
    \label{fig:sessionScatter}
\end{figure}

Among the 108 retained completed sessions, the deployed system, including fusion and gesture-completion logic, achieves 94.4\,\% accuracy, with an APCER of 2.8\,\% and a BPCER of 11.1\,\%~\cite{58:ISO:30107-3:2023}. The observed BPCER indicates that reducing genuine-user rejection remains an important requirement for practical deployment. To compare the aggregate active-capture protocol with its initial calibration segment, we use raw UCF scores from the two segments of the same sessions. The \emph{Hold Still} calibration segment separates attack from genuine sessions with an AUC of 0.878 (95\,\% CI [0.819, 0.963]). The mean across the active-gesture windows reaches 0.988 [0.966, 1.000], a paired difference of $+0.110$ [$+0.032$, $+0.161$]. Restricting each session to its first 5 active actions yields a similar AUC of 0.987 [0.968, 1.000]. Across six leave-one-subject-out analyses, the active-window AUC ranges from 0.983 to 0.998. \paperfigurename~\ref{fig:sessionScatter} shows that several attacks overlap the genuine cluster during calibration but shift upward during the active segment. Thus, under the tested conditions, aggregating responses across randomized actions improves separation without retraining the detector.

\subsection{Gesture-Level Effectiveness}
\label{subsec:PerGesture}

Table~\ref{tab:gestureAUC} compares each gesture category with the \emph{Hold Still} baseline. Pose and complex actions have the highest point estimates against \emph{DeepFaceLive}, but their intervals include zero, and expression shows no clear improvement. The estimates, therefore, do not establish a universal ranking. The session-level evidence is stronger because it aggregates multiple randomized windows.

The clearest category-level difference points downward. Occlusion against \emph{FaceFusion} reaches an AUC of 0.840, compared with 0.910 for \emph{Hold Still}; the paired difference is $-0.069$ [$-0.180$, $-0.017$]. This is exploratory evidence because the intervals are not adjusted for the eight category--generator contrasts. \emph{FaceFusion} was tested with its \texttt{xseg\_1} occlusion mask enabled, so the result is consistent with the mask preserving foreground occluders and reducing visible swap artifacts. However, the experiment did not include a matched run with \texttt{xseg\_1} disabled and therefore cannot isolate its causal effect. Complex actions against \emph{FaceFusion} show the same direction ($-0.043$ [$-0.142$, $+0.024$]), but their interval includes zero. The result illustrates that a challenge reported as effective against earlier pipelines~\cite{65:Mittal:GOTCHA:2024} may provide less evidence against an attack configuration designed to handle it.

\newcommand{\gesturetablecaption}{\caption{Raw UCF AUC by gesture category and paired difference from \emph{Hold Still}. Intervals use a subject-clustered bootstrap and are unadjusted for multiplicity; only the \emph{FaceFusion} occlusion contrast has a 95\,\% interval that excludes zero.}
\label{tab:gestureAUC}}
\begin{paperwidetable}
\ifIEEEpaper\gesturetablecaption\fi
\centering
\footnotesize
\resizebox{0.8\linewidth}{!}{
\begin{tabular}{lcccc}
\toprule
& \multicolumn{2}{c}{\emph{DeepFaceLive}} & \multicolumn{2}{c}{\emph{FaceFusion}} \\
\cmidrule(lr){2-3} \cmidrule(lr){4-5}
\textbf{Category} & \textbf{AUC} & \textbf{$\Delta$ vs.\ baseline [CI]} & \textbf{AUC} & \textbf{$\Delta$ vs.\ baseline [CI]} \\
\midrule
Hold Still (baseline) & 0.847 & --- & 0.910 & --- \\
Expression            & 0.871 & $+0.024$ [$-0.079$, $+0.097$] & 0.924 & $+0.015$ [$-0.054$, $+0.094$] \\
Pose                  & 0.927 & $+0.080$ [$-0.013$, $+0.166$] & 0.943 & $+0.033$ [$-0.036$, $+0.083$] \\
Occlusion             & 0.882 & $+0.035$ [$-0.051$, $+0.085$] & \textbf{0.840} & $\mathbf{-0.069}$ [$-0.180$, $-0.017$] \\
Complex               & 0.932 & $+0.085$ [$-0.002$, $+0.185$] & 0.867 & $-0.043$ [$-0.142$, $+0.024$] \\
\bottomrule
\end{tabular}}
\ifIEEEpaper\else\gesturetablecaption\fi
\end{paperwidetable}

Single-gesture cells are too small for inference; for example, \emph{Smile} against \emph{DeepFaceLive} reaches an AUC of 0.643.

\subsection{Challenge-Response Latency}
\label{subsec:Latency}

Time-to-validation provides a second observable in addition to the detector score. Under the tested conditions, attack sessions take longer to satisfy the validator for all six subjects, with a median within-subject difference of $+1.0$\,s and a range of $+0.2$ to $+1.6$\,s. Pooled over gestures, the median is 3.1\,s for genuine sessions, 4.0\,s for \emph{DeepFaceLive}, and 3.6\,s for \emph{FaceFusion}. Latency alone separates attacks from genuine windows, with an AUC of 0.653 [0.594, 0.734], while the mean latency per session is 0.732 [0.635, 0.870]. These estimates describe the complete attack condition, including synthesis and virtual-camera transport, rather than human response time alone.

\begin{figure}[t]
    \centering
    \includegraphics[width=\latencywidth]{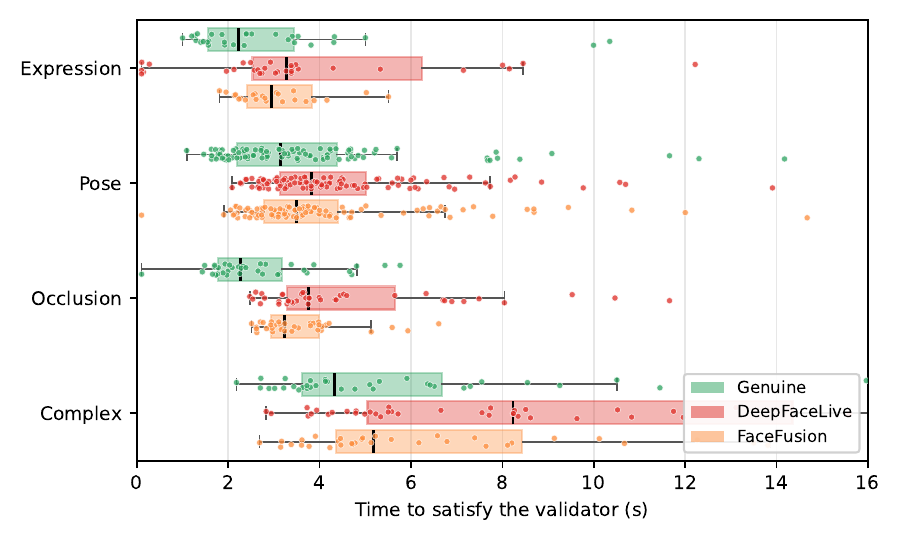}
    \caption{Time to satisfy the challenge validator per gesture category. Attack sessions shift right across all categories, most strongly for occlusion and complex actions.}
    \label{fig:latencyByCategory}
\end{figure}

Occlusion yields the highest latency AUC, 0.800 [0.674, 0.961], with paired differences from pose of $+0.173$ [$+0.097$, $+0.267$] for \emph{DeepFaceLive} and $+0.181$ [$+0.060$, $+0.276$] for \emph{FaceFusion}. Thus, \emph{FaceFusion} occlusion windows combine weaker spatial evidence with a timing difference. Because attacks add synthesis and virtual-camera transport, and timed-out attempts are not retained, latency should be calibrated locally as an end-to-end signal.

\section{Conclusions}
\label{sec:Conclusion}

The static calibration segment reaches a session-level AUC of 0.878 against two unseen attack tools, while the mean across the active-gesture windows reaches 0.988 without retraining the UCF detector. The category results are less definitive: pose and combined tasks have the highest point estimates, expressions show no clear improvement, and face covering against \emph{FaceFusion} scores below the static baseline. Verifier-observed time-to-validation provides additional separation, reaching a session-level AUC of 0.732, although this estimate includes synthesis and transport delay.

Remote verifiers should combine passive inspection with short, randomized actions and re-evaluate these actions as attack tools change. Response time may add evidence, but requires deployment-specific calibration. Within its six-subject, two-tool scope, this proof-of-concept supports active capture as a design principle, not universal performance values or a fixed gesture ranking.

\section*{Acknowledgment}
This work was supported by the Brno University of Technology internal project FIT-S-26-9011. Computational resources were provided by the e-INFRA CZ project (ID:90254), supported by the Ministry of Education, Youth and Sports of the Czech Republic. The authors used ChatGPT and Claude for copy-editing, reviewed all AI-assisted text, and take full responsibility for the content.

\printbibliography
\end{document}